\documentclass[letterpaper, 10 pt, journal, twoside]{IEEEtran}
\usepackage{cite}

\ifCLASSINFOpdf
\else
\fi
\usepackage{amsmath,amssymb,amsfonts}
\usepackage{algorithmic}

\usepackage{array}
\usepackage{tabularx}
\usepackage{booktabs}
\usepackage{multirow}
\usepackage{hhline}

\usepackage[caption=false,font=footnotesize]{subfig}
\usepackage{graphicx}
\usepackage{textcomp}
\usepackage{xcolor}
\usepackage{bbm}
\usepackage{dsfont}
\usepackage{hyperref}

\newcommand{\squeezeup}{\vspace{-2.5mm}}
\usepackage{epstopdf}
\AppendGraphicsExtensions{.tiff}

\usepackage{url}

\newcommand{\norm}[1]{\left\lVert#1\right\rVert}
\DeclareMathOperator{\sech}{sech}

\begin{document}
%
\title{Learning and Leveraging Features in Flow-Like Environments to Improve Situational Awareness}
%
%
%

\author{Tahiya Salam$^{1}$, Victoria Edwards$^{1}$, and M. Ani Hsieh$^{1}$%
\thanks{Manuscript received: September 9, 2021; Revised December 3, 2021; Accepted December 30, 2021.}
\thanks{This paper was recommended for publication by Editor Youngjin Choi upon evaluation of the Associate Editor and Reviewers' comments.
This work was supported by ARL DCIST CRA W911NF-17-2-0181 and NSF IIS 1812319.} 
\thanks{$^{1}$All authors are with the General Robotics Actuation Sensing
and Perception (GRASP) Laboratory at the University of Pennsylvania, Philadelphia, PA, USA
        {\tt\footnotesize \{tsalam, vmedw, m.hsieh\}@seas.upenn.edu}}%
\thanks{Digital Object Identifier (DOI): see top of this page.}
}
%
%

\markboth{IEEE Robotics and Automation Letters. Preprint Version. Accepted January, 2022}
{Salam \MakeLowercase{\textit{et al.}}: Features in Flow-Like Environments} 

%



\maketitle

\begin{abstract}
This paper studies how global dynamics and knowledge of high-level features can inform decision-making for robots in flow-like environments. Specifically, we investigate how coherent sets, an environmental feature found in these environments, inform robot awareness within these scenarios. The proposed approach is an online environmental feature generator which can be used for robot reasoning. We compute coherent sets online with techniques from machine learning and design frameworks for robot behavior that leverage coherent set features. We demonstrate the effectiveness of online methods over offline methods. Notably, we apply these online methods for robot monitoring of pedestrian behaviors and robot navigation through water. Environmental features such as coherent sets provide rich context to robots for smarter, more efficient behavior. \end{abstract}


%
\IEEEpeerreviewmaketitle

\section{Introduction}
%
%
%
%
\IEEEPARstart{R}{obots} operating in flow-like environments benefit from a greater understanding of the characteristics of their surroundings. Flow-like environments, such as ocean and air currents, movements of vehicles and pedestrians, or transport of blood in the heart, are often complex and hard to model. However, robots conducting environmental monitoring, autonomous driving, crowd monitoring, or surgical procedures all require knowledge of these flow-like environments in their tasks. The ability to deduce and incorporate the important features of these environments into planning and decision-making allows robots to operate more successfully. We propose a strategy for robots to learn high-level features in flow-like environments and show how these features can be incorporated into existing planning frameworks. 

\begin{figure}[ht!]
    \captionsetup[subfloat]{farskip=2pt,captionskip=1pt}
    \centering
    \subfloat[ ]{\includegraphics[width=0.46\textwidth]{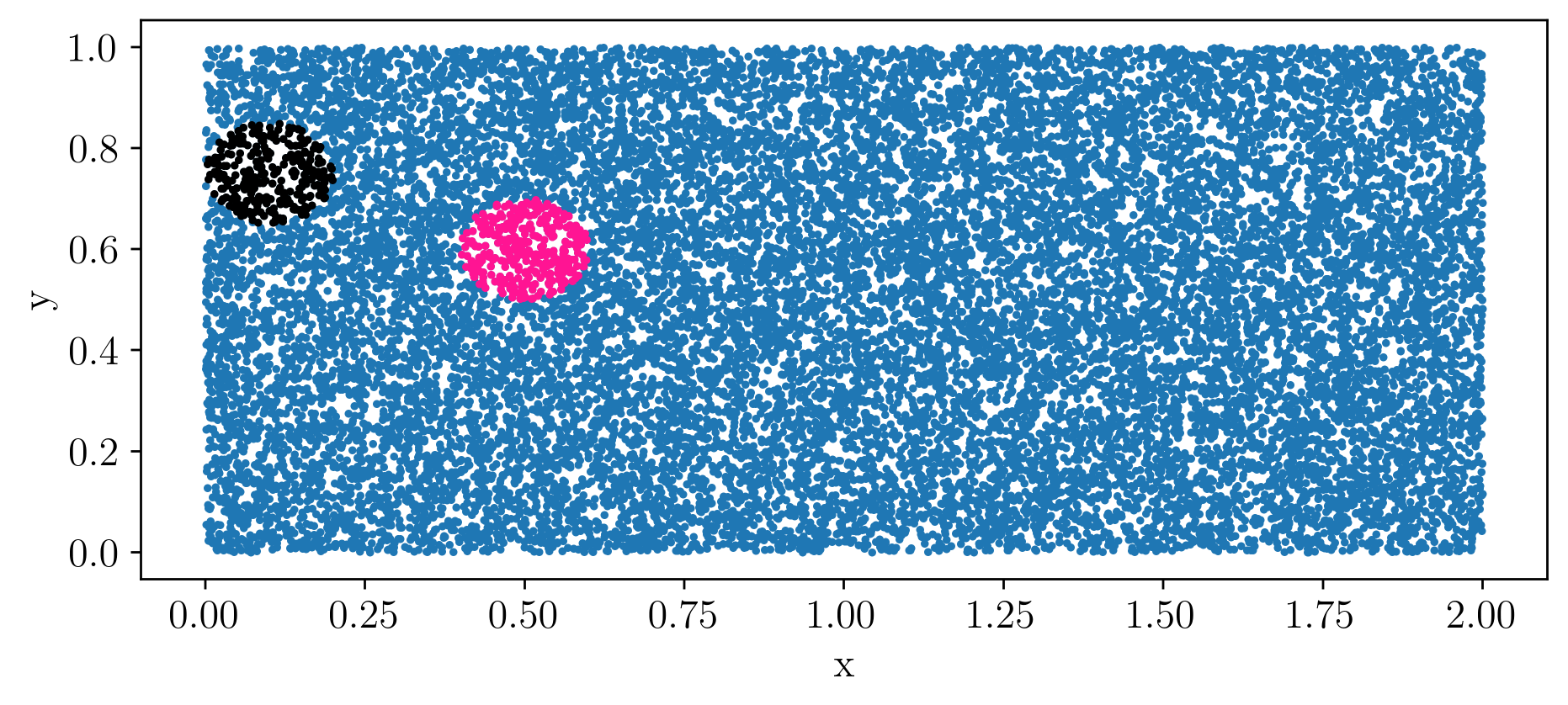}}

    \subfloat[ ]{\includegraphics[width=0.46\textwidth]{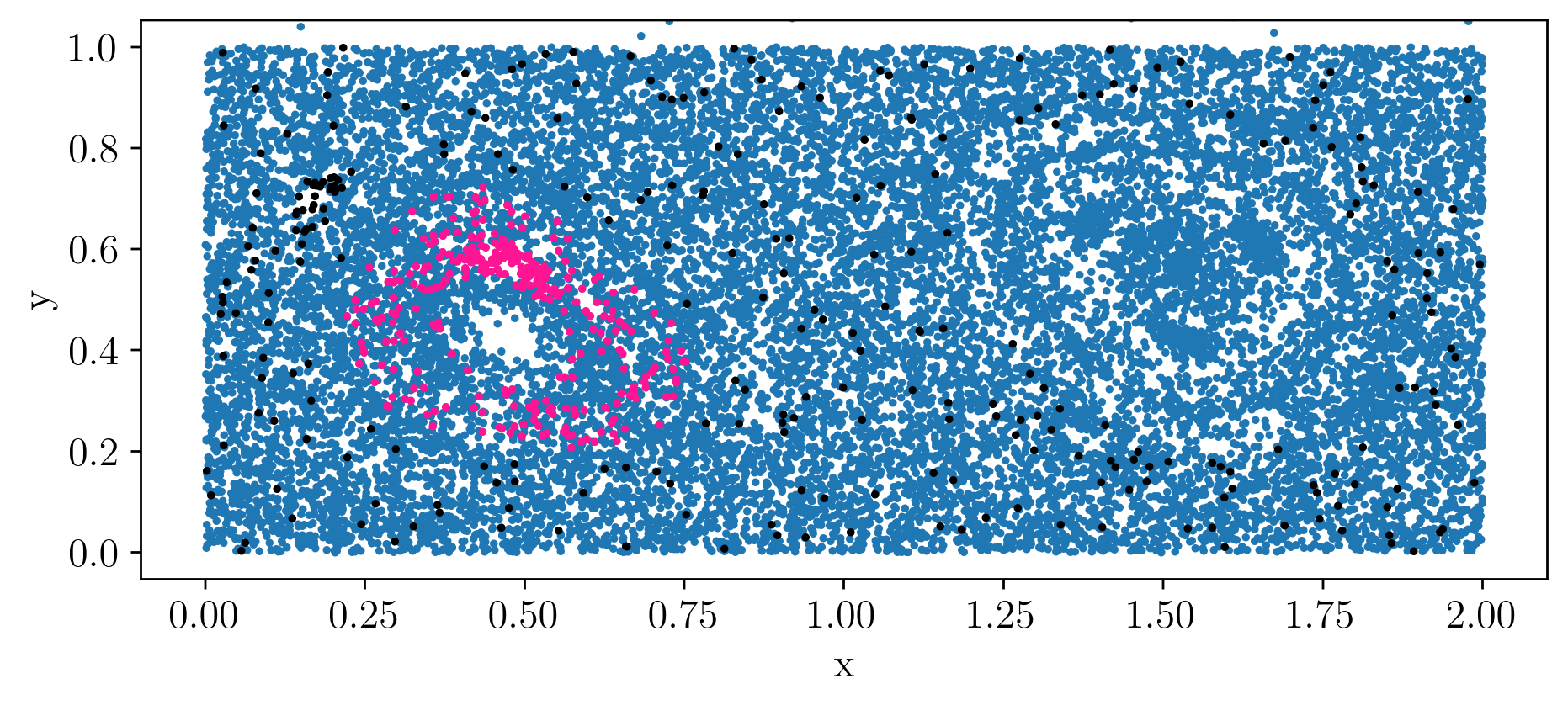}}
    \caption{Time-dependent double gyre flow at times (a) $t = 0$ and (b) $t = 5.1$. The double gyre model is a prototypical model in the study of coherent sets. The black particles, not in a region of coherence, disperse quickly. The pink particles within a coherent structure follow the flow and stay within a region.} \label{fig:coherence-dg}
    \squeezeup
\end{figure}

The typical study of flow-like environments requires knowledge of the flow map or vector field that describes the evolution of the entire state of the system under a specified time frame. Robotics application within these environments usually involve detailed knowledge and representation of the underlying flow velocity \cite{Kularatne2018OptimalDiscretization, Lolla2012PathMethods, Palmieri2017KinodynamicFields}, which is not always feasible and may be computationally expensive. The methods in this paper allows us to represent key features of the flow-like environment without explicitly estimating or tracking the equations of motions. Here, we rely on the extraction of the global dynamics of the system, namely coherent sets, from noisy data to capture high-level, overarching spatial trends of the particles. 

Coherent sets are a global characteristic of turbulent flows that describe groups of particles that move together robustly and do not disperse much \cite{Fiedler1988CoherentFlows}. This means particles within coherent sets tend to stay within the sets (see Fig. \ref{fig:coherence-dg}). Lagrangian coherent structures (LCS) are separatices that delineate dynamically distinct regions in dynamical systems \cite{Kelley2013LagrangianFlows}. LCS provide different but complementary information when compared to coherent sets \cite{Tallapragada2013ASets}. Knowledge of coherent sets and LCS allows robots to compactly represent key characteristics of these flow-like environments and leverage them in application-specific ways. For example, surgical robots could understand vasculature conditions such as flow stagnation and separation during medical procedures \cite{Shadden2008CharacterizationSystem}. Aerial vehicles are affected by coherent features during takeoff and landing \cite{Tang2011LagrangianData}. Robots used for environmental monitoring may be interested in atmospheric and oceanic transport pathways that play a role in the movement of pollutants \cite{Nolan2020PollutionStructures}. Marine vehicles can plan energy efficient trajectories in the ocean by leveraging coherent structures \cite{Inanc2005OptimalFlows, Heckman2016ControllingEnvironments, Ramos2018LagrangianMissions} and maintaining sensors in their desired monitoring regions \cite{Hsieh2014DistributedFlows, Wei2019Low-rangeStructures}. Knowledge of the coherent sets improves robots' situational awareness in context-specific ways by offering a lightweight feature of the complex environment that can be used for decision-making and high-level, coarse planning.

There are many methods for studying coherent sets through transfer operators which provide a framework for analyzing the global behaviors of dynamical systems. These methods identify which geometric structures are the strongest barriers to global mixing \cite{Dellnitz1999OnBehavior, FROYLAND_2009, Froyland2010CoherentSystems, Froyland2010TransportSetsb}. There are also methods for identifying LCS relying on detailed flow field data and integration over a parametrized time interval \cite{Haller2002LagrangianData}. While these approaches elucidate coherence that can be used for robot planning and control, none of these methods take into account the complex interdependencies within the data sets and require detailed environmental knowledge. Additionally, for many robotic applications, online algorithms are preferred, especially when the time parameterization of the system is unknown in the absence of prior models or historical data. 

In recent years, there has been a surge of machine learning approaches relying on kernels (or feature maps) and transfer operators to estimate dynamical systems and their key characteristics \cite{Williams2015AAnalysis, Hamzi2021Data-drivenMethods, Bouvrie2017KernelSystems, Klus2020EigendecompositionsSpaces, Klus2019KernelData, Takeishi2019KernelOperators, Klus2020Kernel-basedOperator}. Kernels map the environmental data to a higher dimensional feature space with greater expressivity for learning models. New advances in the detection of coherent sets leverage techniques from machine learning \cite{Klus2020EigendecompositionsSpaces, Klus2019KernelData}. Though these techniques highlight the expressiveness of data-driven, kernel-based methods, they still suffer from the same limitations in their assumptions about centralized, offline data, and time parameterization. 

Despite the field's growing interest in transfer operators \cite{Abraham2017Model-BasedOperators, Abraham2019ActiveOperators, Bruder2019ModelingControl, Folkestad2020EpisodicLanding}, to the best of our knowledge, these advances at the intersection of machine learning and dynamical systems have not been explored in a robotics context using online algorithms. Our work builds upon these recent advances to develop strategies that allow robots to compute coherent set features online and contextualize them in the robots’ environment for increased situational awareness and to facilitate decision-making. The contributions of this paper are: 
\begin{itemize}
    \item online feature detection methods within complex environmental structures and
    \item demonstrations of how coherent set features can be used for decision-making and planning in two application settings.
\end{itemize}

We propose using the coherent set detection framework as a map for greater situational awareness for operating within flow-like environments that exhibit this notion of coherence. First, we provide the relevant background in dynamical systems and machine learning related to coherent sets and show how kernel methods can represent coherent sets from data. Next, we demonstrate the usefulness of computing coherent sets in an online setting with simulations and analysis of known benchmarks. We apply these online methods for computing coherent sets for overhead data in an urban monitoring environment, where the coherent sets correspond to regions of interest for situational awareness. Lastly, we show how sparse representations of coherent sets in a fluid flow can be leveraged for energy-efficient navigation of surface vehicles. 

\section{Methodology}
\begin{figure*}
    \squeezeup
    \includegraphics[width=\textwidth]{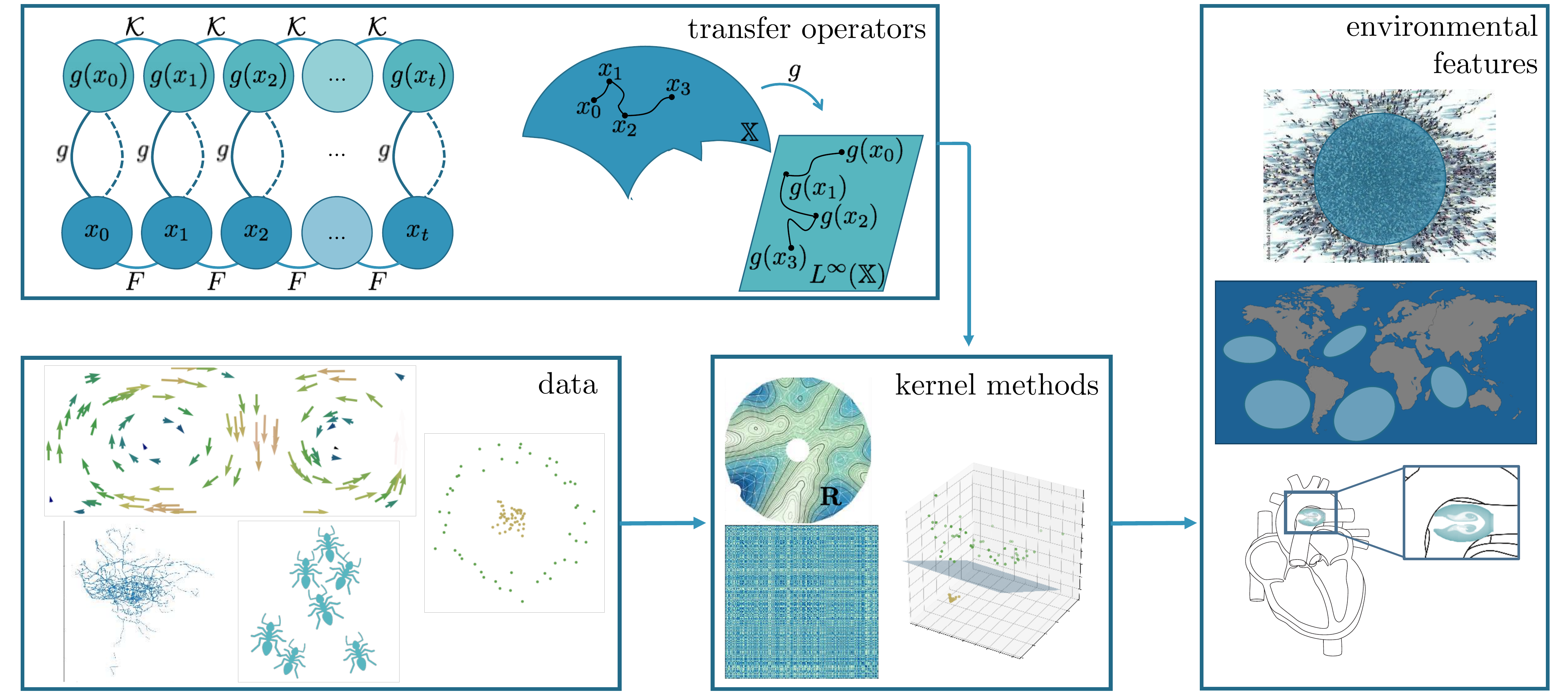}
    \caption{Diagram of interplay between data, transfer operators, kernel methods, and environmental features. Transfer operators represent dynamical systems, where a state $x \in \mathbb{X}$ is lifted to a space $L^{\infty}\mathbb{X}$ and $g(x)$ provides physical properties of the system. Many systems are defined by data exhibiting complex patterns, such as two nested rings, flows in oceans, taxi trajectories, and biological behaviors. Kernel methods transform this data to an alternative space with the use of kernel functions. Data is then easier to interpret, such as by separating two nested rings or by creating a Gram matrix for use in a kernel algorithm. Transfer operators are represented through kernel methods by embedding dynamical systems into a kernel space, $\mathbf{R}$. Kernel algorithms extract environmental features from transfer operators, such as where humans tend to congregate in crowds, areas of gyres in oceans, or patterns of blood flow in hearts.}
    \squeezeup
    \label{fig:connections}
\end{figure*}

\subsection{Dynamical Systems and Transfer Operators}
Coherent sets are global properties of a dynamical system,  extracted through constructing transfer operators. In this section, we discuss the connection between dynamical systems and transfer operators.

Define a discrete dynamical system as $ x_{i+1} = F(x_{i})$ where $F: X \rightarrow X$ is a measurable map. Transfer operators allow for a systematic study of the global properties of dynamical systems and alleviates computing the map $F$, which is often difficult to model and estimate for unknown environments exhibiting complex, nonlinear dynamics. For example, the Koopman operator is an infinite dimensional linear transfer operator that fully captures the nonlinear dynamics of the system by acting on the space of scalar-valued functions operating on the observables of the system, $g(x)$. 

Transfer operators can more generally be defined for stochastic dynamical systems to infer statistical aspects of a dynamical system. Define $\{X_{t}\}_{t \geq 0}$ as a stochastic process on the state space $\mathbb{X} \in \mathbb{R}^{d}$. $\{X_{t}\}_{t \geq 0}$ is a stationary and ergodic Markov decision process. A transition density function, $p_{\tau}$, is then defined by the relation $\mathbb{P}[X_{t+\tau} \in \mathbb{A} | X_{t} = x] = \int_{\mathbb{A}} p_{\tau}(y|x) dx$ for some measurable set $\mathbb{A}$.

Given a probability density $p_t \in L^{1}(\mathbb{X})$, an observable of the system $f_{t} \in L^{\infty}(\mathbb{X})$, and a specified lag time $\tau$, we can define the following transfer operators that provides information about iterated maps. Specifically, the Perron-Frobenius and Koopman transfer operators serve as adjoints to one another. Furthermore, the eigenvalues of transfer operators are real and their eigenfunctions form an orthogonal basis with respect to the corresponding scalar product. The Perron-Frobenius, $\mathcal{P}$, and Koopman, $\mathcal{K}$, transfer operators are defined as follows: $\mathcal{P} : L^{1}(\mathbb{X}) \rightarrow L^{1}(\mathbb{X})$ where $\mathcal{P}p_{t}(y) = \int p_{\tau}(y|x) p_{t}(x) dx$, while $\mathcal{K} : L^{\infty}(\mathbb{X}) \rightarrow L^{\infty}(\mathbb{X})$ where $\mathcal{K}f_{t}(x) = \int p_{\tau}(y|x)f_{t}(y)dy = \mathbb{E}[f_{t}(X_{t+\tau}) | X_{t} = x]$.

Transfer operators and their decompositions are often approximated using methods acting directly on observations of the dynamical systems, {\it i.e.}, data \cite{Mezic2005, Klus2020EigendecompositionsSpaces, Das2020KoopmanSpaces, Kawahara2016DynamicAnalysis}.
\subsection{Kernel Methods}
Instead of constructing models using data directly, kernel methods rely on transforming data into a feature representation that elucidates interesting, meaningful patterns. Here, we present the relevant background for kernel methods, which later serves as the foundation for estimating transfer operators.

Kernel functions $k: X \times X \rightarrow \mathbf{R}$ are applied to elements of some space, $X$, to measure the similarity between any pairs of elements. However, this similarity metric can be found by computing the inner product of features $\phi(x)$ acting on elements of the input space in some high-dimensional, possibly infinite feature space in $\mathbf{R}^N$. Thus, kernels allow us to compare objects based on their features. Formally, for inputs $x, x' \in X$, a feature map $\phi: X \rightarrow \mathbf{R}^{N}$, and some valid inner product $\langle\cdot{ , }\cdot\rangle_\mathcal{V}$, a kernel $k$ is defined as 
\begin{equation}\label{kernel-trick}
    k(x, x') = \langle\,\phi(x), \phi(x')\rangle_\mathcal{V}.
\squeezeup
\end{equation} 
By using the relation in Eq. \eqref{kernel-trick}, commonly referred to as the kernel trick, operations in the high-dimensional feature space can be computed without ever explicitly computing the coordinates of $\phi(x)$ and instead using an efficient kernel function in inner product space \cite{Hofmann2008}. 

Kernels on features spaces that are positive definite can be used to define a function $f$ on $X$. The space of such functions $f$, such that the evaluation of $f$ at $x$ can be represented as the inner product $\langle\,\phi(x), \phi(x')\rangle$ in feature space, is referred to as the Reproducing Kernel Hilbert Space (RKHS). An RKHS is a space $H$ of functions mapping $X$ to $\mathbf{R}$ with two defining features:
\begin{gather} 
\forall x \in X, \phi(x) \in H \label{feature}, \\
\forall x \in X, \forall f \in H \phi(x) \in H, \langle\,f, \phi(x)\rangle_{H} = f(x). \label{reproducing}
\end{gather}

These two properties show that in an RKHS the feature map of every point is in the feature space as in \eqref{feature} and that the space exhibits the reproducing property as in \eqref{reproducing}. The reproducing property defines a functional evaluation at a point as the inner product between the function and evaluated feature map \cite{Gretton2013IntroductionAlgorithms}.

Let $\phi$ be the feature map associated with kernel $k$ defined on $X$ and $\psi$ be the feature map associated with kernel $l$ defined on $Y$. Then, define feature matrices as $\Phi = [\phi(x_1) \dots \phi(x_n)]$ and $\Psi = [\psi(y_1) \dots \psi(y_n)]$, with Gram matrices $G_{XX} = \Phi^{T} \Phi$ and $G_{YY} = \Psi^{T}\Psi$. $C_{XX}$ is then the covariance operator and $C_{YX}$ is the cross-covariance operator, which can be thought of as a nonlinear generalization of standard covariance and cross-covariance matrices \cite{Klus2020EigendecompositionsSpaces}. $C_{XX}$ and $C_{YX}$ can be estimated as  
\begin{gather} 
    \hat{C}_{XX} = \frac{1}{n}\Phi\Phi^{T} \label{c-xx-hat}, \\
    \hat{C}_{YX} = \frac{1}{n}\Psi\Phi^{T} \label{c-yx-hat}.
\end{gather}

The covariance and cross-covariance operators are one of the most important and established tools in RKHS theory. These operators are used in the computation of kernel principal components analysis (PCA), the kernel Fisher discriminant, kernel partial least squares, the kernel canonical correlation analysis (CCA), and many other learning algorithms \cite{Gretton2015NotesOperators}. Some learning algorithms relying on the use of (\ref{c-xx-hat}) and (\ref{c-yx-hat}), can be written in terms of their corresponding Gram matrices. For Gram matrices, only the inner products between $\phi(x)$ and $\phi(x')$ and $\psi(x)$ and $\psi(x')$ need to be computed. Thus, by computing Gram matrices on a finite number of data points, we can alleviate the need to explicitly compute the possibly infinite-dimensional covariance and cross-covariance operator. The ability to construct algorithms in the inner product space using kernels, and more specifically the kernel trick, lends them to be particularly useful in a variety of learning scenarios \cite{Hofmann2008}. In the next section, we will lay the framework for use of transfer operator theory in the RKHS and later show how this expression in RKHS can be used as a learning technique for modeling coherent sets and decision-making with this awareness.

\subsection{Connecting Transfer Operators to Kernel Methods}
Transfer operator approximation in the RKHS allows for the use of kernel-based methods, acting on feature maps of the data, while still capturing the properties of the transfer operators. Transfer operators can be expressed in terms of the covariance and cross-covariance operators defined on some RKHS \cite{Klus2020EigendecompositionsSpaces}. Assume the input and output space are the same, {\it i.e.}, $X = Y$, then the kernels and resulting Hilbert spaces are also the same. In this scenario, we assume $g(x) = x$, meaning full-state observables are simply the states themselves.

The kernel Koopman operator is defined as 
\begin{equation}\label{kernel-koopman}
    \mathcal{K}_k f(x) = \langle C_{XX}^{-1} C_{XY} f \, , k(x,\cdot) \rangle 
\end{equation}
and pushes forward the observables. Using the approximation of $C_{XX}$ from (\ref{c-xx-hat}) and $C_{XY}$ from the transpose of (\ref{c-yx-hat}), the empirical estimate of the Koopman operator is 
\begin{equation}\label{kernel-koopman-est}
    \hat{\mathcal{K}}_k = C_{XX}^{-1} C_{XY} = (\Phi \Phi^{T})^{-1}(\Phi \Psi^{T}) = \Phi G_{XX}^{-1} \Psi^{T}. 
\end{equation}
This value can be calculated empirically from training data.  Similarly, the kernel Perron-Frobenius operator is defined as 
\begin{equation}\label{kernel-perron-frobenius}
    \mathcal{P}_k p(x) = \langle C_{XX}^{-1} C_{YX} p \, , k(x,\cdot) \rangle
\end{equation}
with the empirical estimate of the Perron-Frobenius as
\begin{equation}\label{kernel-perron-frobenius-est}
    \hat{\mathcal{P}}_k = C_{XX}^{-1} C_{YX} = (\Phi \Phi^{T})^{-1}(\Psi \Phi^{T}) = \Psi A \Phi^{T} 
\end{equation}
with $A = G_{XY}^{-1} G_{XX}^{-1} G_{XY}$. For complete derivations, see  \cite{Klus2020EigendecompositionsSpaces}.

We now have a representation of the transfer operators in terms of the covariance and cross-covariance operators that can be estimated using the kernel versions \eqref{c-xx-hat} and \eqref{c-yx-hat}. Thus, we now have a framework unifying existing techniques for approximating transfer operators and their eigendecompositions in RKHS \cite{Klus2020EigendecompositionsSpaces}, as shown in Fig. \ref{fig:connections}. In our work, we can use additional techniques from statistics to extract the dominant characteristics from the transfer operator representation of the environment, specifically coherent sets.


\subsection{Kernel Methods for Coherent Set Detection}
Here, we will provide the mathematical definition of coherent sets, show their representation through transfer operators, and demonstrate how this transfer operator representation can be computed using their kernel versions. We refer the reader to \cite{Klus2019KernelData} for a detailed discussion on these methods and the intuition behind how these methods relate to coherence. Let $\mathcal{F} : L_{\mu}^2(\mathbb{X}) \rightarrow L^2(\mathbb{Y})$ be the forward operator, for some $\mu$ that is a reference density of interest. Given $\nu = \mathcal{F} \mathds{1}$ as the image density obtained by mapping the indicator function on $\mathbb{X}$ forward in time. If we normalize $\mathcal{F}$ with respect to $\nu$, we can define a new operator $\mathcal{A}: L^2_{\mu}(\mathbb{X}) \rightarrow L^2_{\nu}(\mathbb{Y})$ and its adjoint $\mathcal{A^*}: L^2_{\nu}(\mathbb{Y}) \rightarrow L^2_{\mu}(\mathbb{X})$ as
\begin{gather*}\label{a-adjoint}
\squeezeup
\squeezeup
    (\mathcal{A}f)(y) = \int{\frac{p_{\tau}(y|x)}{\nu(y)}}f(x) \mu(x) dx, \\
    (\mathcal{A}^*g)(x) = \int{p_{\tau}(y|x) g(y) dx}.
\end{gather*}

Here, $\mathcal{A}$ is the analogue of the reweighted Perron-Frobenius operator, or the Koopman operator with the roles of $X$ and $Y$ reversed, and $\mathcal{A}^*$ is the analogue of the Koopman operator. \textbf{The operator $\mathcal{A^{*} \mathcal{A}}$ is used to detect coherent sets.} To detect the coherent sets, the following eigenvalue problems needs to be solved:
\begin{equation}
    \mathcal{A}^*\mathcal{A}f = \rho^2 f \label{astar-soln}.
\end{equation}



In general, the inverses of the covariance and cross covariance operators do not exist. Thus, the regularized inverses are used, specifically we use $(C_{XX} + \epsilon \mathcal{I})^{-1}$ and $(C_{YY} + \epsilon \mathcal{I})^{-1}$. We can then define the RKHS approximation of $\mathcal{A}$ as $(C_{XX} + \epsilon \mathcal{I})^{-1}C_{XY}$ and $\mathcal{A}^*$ as $(C_{YY} + \epsilon \mathcal{I})^{-1}C_{YX}$ from the kernel formulations of transfer operators in \eqref{kernel-koopman} and \eqref{kernel-perron-frobenius}.

Using these regularized inverses to solve (\ref{astar-soln}) gives 
\begin{equation}\label{kernel-CCA-soln-inv}
    (C_{XX} + \epsilon \mathcal{I})^{-1} C_{XY} (C_{YY} + \epsilon \mathcal{I})^{-1} C_{YX} f = \rho^2 f.
\end{equation}


Using the empirical estimates from \eqref{kernel-koopman-est} and \eqref{kernel-perron-frobenius-est}, the eigenvalue problem of the RKHS version of the transfer operators in Eq. \eqref{kernel-CCA-soln-inv} becomes
\begin{equation*}\label{kernel-CCA-soln-inv-est}
    \squeezeup
    \Phi B \Phi^T \hat{f} = \rho^2 \hat{f} 
\end{equation*} 
for $B = (G_{XX} + n\epsilon I)^{-1} (G_{YY} + n\epsilon I)^{-1} G_{YY}$. Instead of solving this eigenvalue problem directly, we can instead solve a surrogate eigenvalue problem
\begin{equation}\label{surrogate-eig-val-prob}
\begin{split}
\squeezeup
\squeezeup
        (G_{XX} + n\epsilon I)^{-1}(G_{YY} + n\epsilon I)^{-1}G_{YY}G_{XX}v &= \rho^2 v \text{, } \\ \text{with } \hat{f} &= \Phi v.
\end{split}
\end{equation}
In practice, we will solve this eigenvalue problem to compute the coherent sets of a system. In order to compute the coherent sets, we first compute the dominant eigenfunctions, as in Eq. \eqref{surrogate-eig-val-prob}. Then, a $k$-means clustering is applied to the $k$ dominant eigenfunctions to determine the coherent sets. 
\squeezeup
\subsection{Online Computation of Coherent Sets}\label{sec:online}
Robots collect data in real time and knowledge of the dynamics of the environment is limited; thus, we desire a way to detect coherent sets online using $\mathcal{A}^*\mathcal{A}$ and solving the eigenvalue problem, estimated as Eq. \eqref{astar-soln}. The operator $\mathcal{A}^*\mathcal{A}$ can be represented as a function of covariance operators, $\mathcal{C}_{XX}$ and $\mathcal{C}_{YY}$, and cross covariance operators, $\mathcal{C}_{XY}$ and $\mathcal{C}_{YX}$. Instead of using covariance and cross covariance operators directly, the eigenvalue problem is solved using a surrogate formulation \eqref{surrogate-eig-val-prob}. The surrogate formulation relies on Gram matrices $G_{XX}$ and $G_{YY}$, where $G_{XX}$ captures correlations between the initial positions of the tracked objects and $G_{YY}$ captures correlations between the final positions of the objects. However, robots collect sensor measurements in real time. Other works assume knowledge of the fixed lag time $\tau$, such that analyzing the operator $\mathcal{A}^*\mathcal{A}$ with data at $t_0$ and $\tau$ computes coherent sets \cite{Klus2019KernelData}. Our paper analyzes how coherence can be computed in an online setting with no prior assumptions about $\tau$. In this section, we extend the work from \cite{Klus2019KernelData} and use insights in dynamical systems theory from \cite{Froyland2010TransportSetsb, Froyland2013AnSystems, Banisch2017UnderstandingSets} to offer a mathematically sound method for computing coherent sets online using kernel-based techniques from machine learning.

For coherence at several time instances, studies have shown it is sufficient to average the operators $\mathcal{A}^*\mathcal{A}$ for all the different time instances and compute the dominant eigenfunctions of the resulting operator \cite{Froyland2010TransportSetsb, Froyland2013AnSystems, Banisch2017UnderstandingSets}. Let $\mathcal{A}^*_{t_0, t}$ correspond to the Koopman operator and $\mathcal{A}_{t_0, t}$ correspond to the Perron-Frobenius operator for the dynamics from $t_0$ to $t$.  Then, we are interested in the time-averaged quantity  
\begin{equation} \label{time-avg-astar-a}
        \frac{1}{T}\sum\limits_{i = 0}^{T-1} \mathcal{A}^*_{0, i}\mathcal{A}_{0, i}.
\end{equation}

Averaging the operator $\mathcal{A}^*\mathcal{A}$ by using the representations of the operators in \eqref{kernel-CCA-soln-inv} and the corresponding surrogate eigenvalue problem in \eqref{surrogate-eig-val-prob}, we are interested, at each time step $t$, the eigenvalue decomposition of   
\begin{equation*}
    \squeezeup
    \frac{1}{t}\sum\limits_{i=0}^{t} (G_{XX} + n\epsilon I)^{-1}(G_{Y_iY_i} + n\epsilon I)^{-1}G_{Y_iY_i}G_{XX},
\end{equation*}
which can be rewritten as 
\begin{equation} \label{time-dep-kcca}
    \squeezeup
    (G_{XX} + n\epsilon I)^{-1} \left[ \frac{1}{t}\sum\limits_{i=0}^{t-1} (G_{Y_iY_i} + n\epsilon I)^{-1} G_{Y_iY_i}\right] G_{XX}. 
\end{equation}
The inner sum is computed as the standard optimal online averaging solution with the addition of more data. Then, the eigenvalue decomposition of the full resulting operator for the surrogate problem, Eq. \eqref{time-dep-kcca}, is computed at each time step. Finally, $k$-means clustering is applied, as before.

\section{Simulations and Analysis}\label{sec:sim-and-analysis}
We present two examples of fluid flows, where the coherent sets are well-studied. We perform an analysis on the quality of the features generated from our online method compared to those from offline methods. We use known systems to analyze the performance of our online methods since it is hard and often infeasible to obtain ground truth for coherent sets for real-world applications, especially when the underlying dynamics are poorly understood.
\subsection{Time-Dependent Double Gyre} \label{subsec:double-gyre}

A simple model of the wind-driven, time-dependent double gyre flow, as in \cite{Forgoston2011Set-basedInvariant}, is described by
\begin{align*}
    f(x, t) &= \epsilon * \sin{(\omega t)} * x^2 + (1 - 2 \epsilon \sin{(\omega t})) * x \\
    \frac{\partial f}{\partial x} &= 2 \alpha * \sin{(\omega t)} * x + (1 - 2 \alpha \sin{(\omega t})) \\
    \dot{x} &= -\pi A \sin{(\pi f(x,t))} \cos{(\pi y)} \\
    \dot{y} &= \pi A \cos{(\pi f(x,t))} \sin{(\pi y)} * \frac{\partial f}{\partial x}.
\end{align*}
\begin{figure}[!hb]
    \squeezeup
    \captionsetup[subfloat]{farskip=2pt,captionskip=1pt}
    \centering
    \subfloat[ ]{\includegraphics[width=0.48\textwidth]{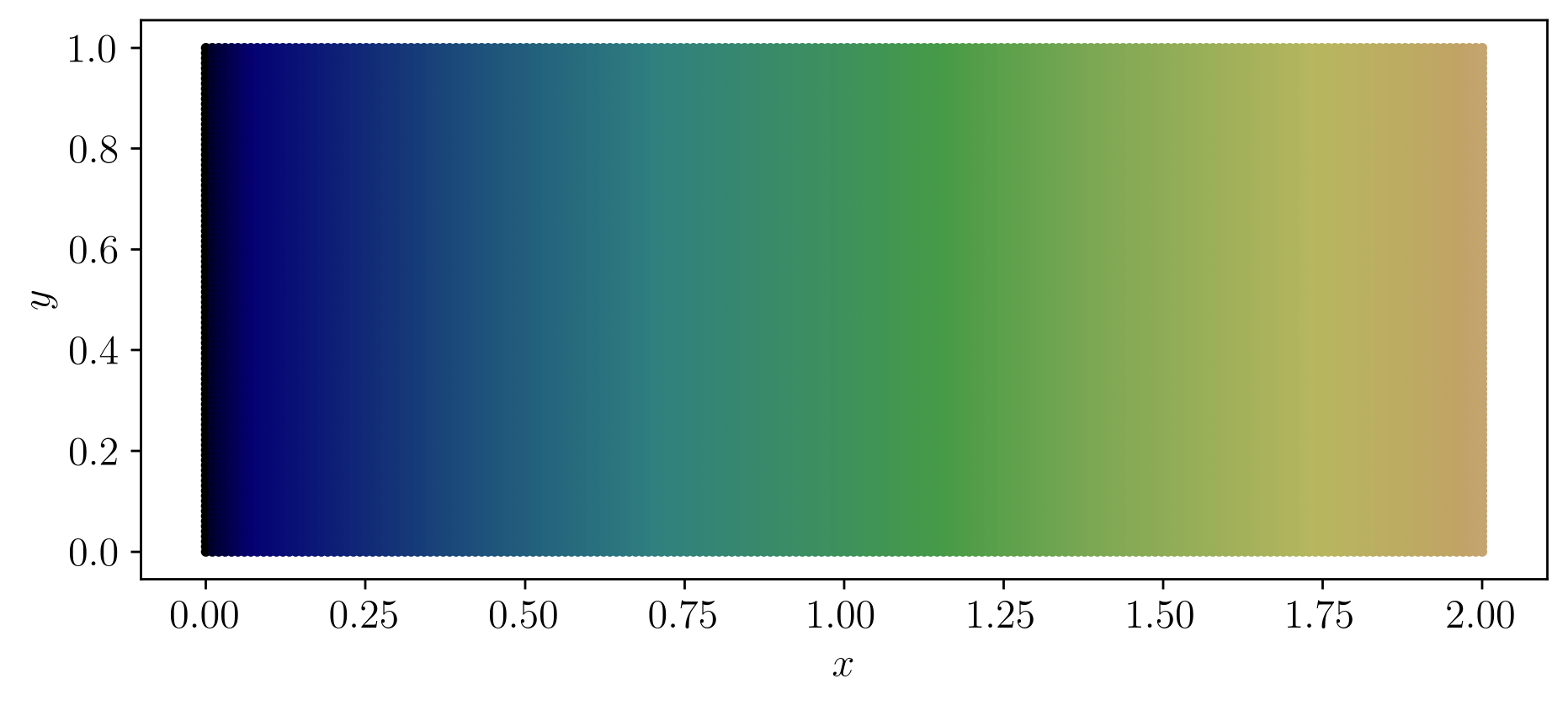}}
    
    \subfloat[ ]{\includegraphics[width=0.48\textwidth]{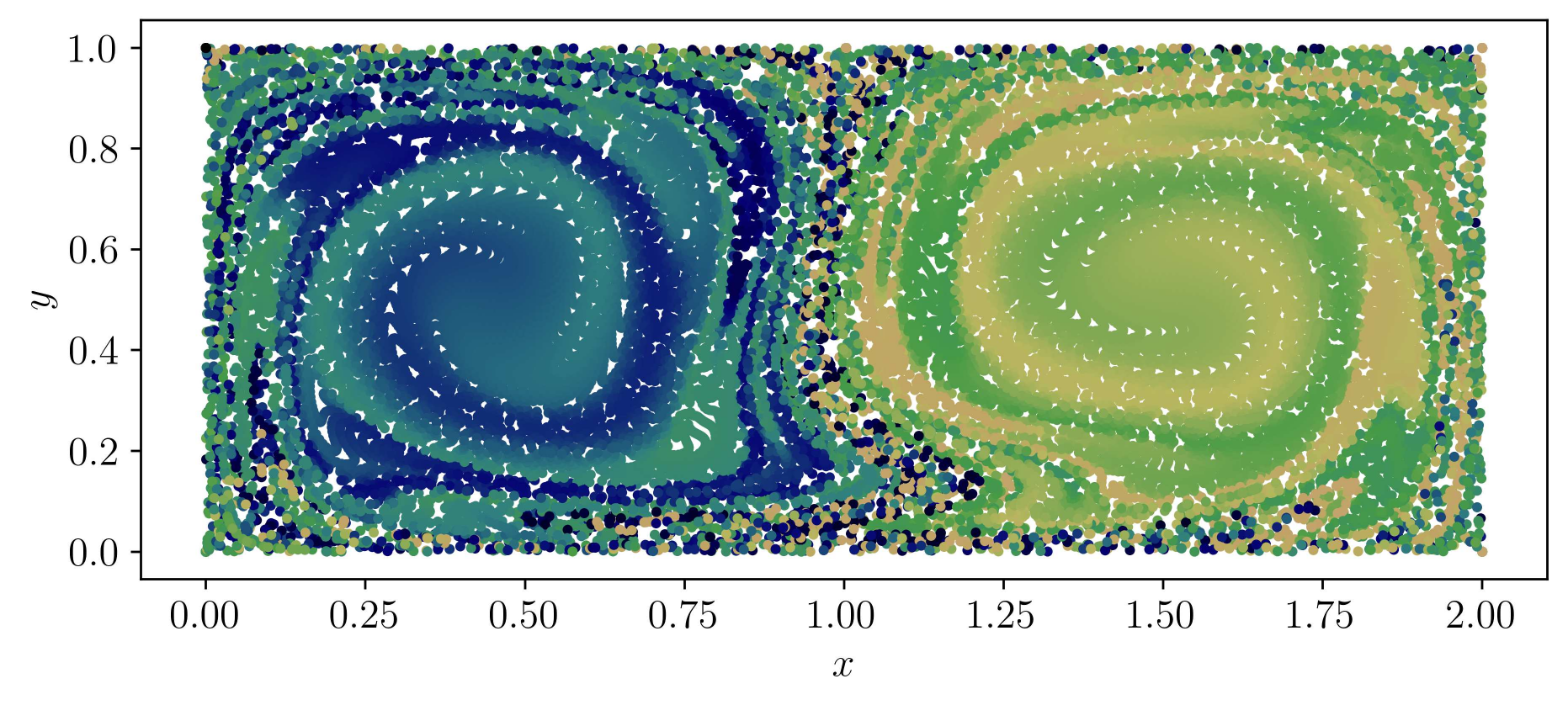}}
    \caption{(a) $20\,000$ particles are initialized in a $[0, 2] \times [0, 1]$ environment. Particles are colored by their $x$-coordinate to visualize their movement. (b) The position of the particles are tracked under a time-dependent double gyre flow at time $t = 12.1$. Two dynamically distinct regions can be seen as a result of this type of flow, which correspond to coherent sets.} \label{fig:double-gyre-overview}
    \squeezeup
\end{figure}

\begin{figure}[!ht]
    \squeezeup
    \captionsetup[subfloat]{farskip=2pt,captionskip=1pt}
    \centering
    \subfloat[ ]{\includegraphics[width=0.48\textwidth]{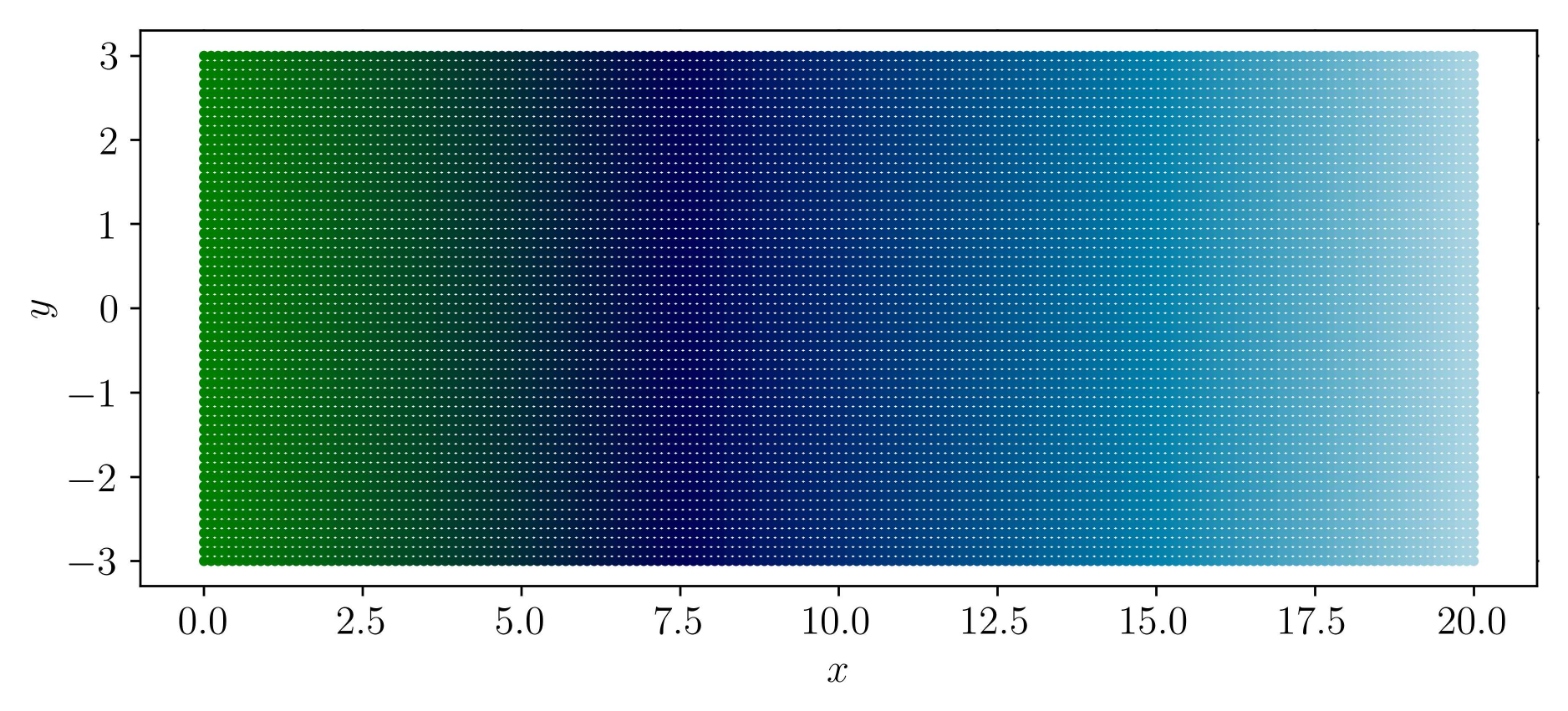}}
    
    \subfloat[ ]{\includegraphics[width=0.48\textwidth]{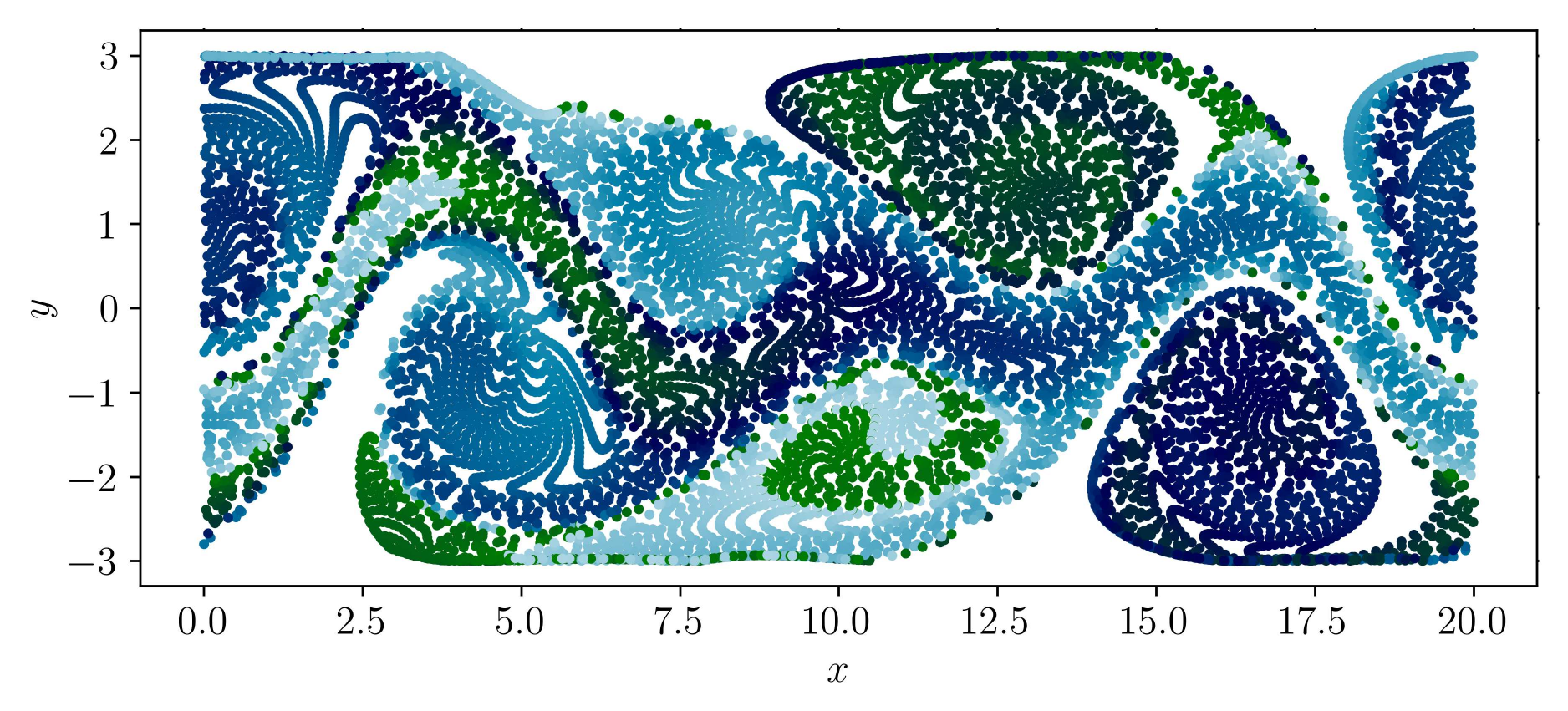}}
    \caption{The Bickley jet is an idealized model for fluid flows, such as in the ocean or in the atmosphere. (a) $9900$ particles are initialized in a $[0, 20] \times [-3, 3]$ environment. (b) The movement of the particles are tracked under a Bickley jet model until $t=40$, shown here at $t = 10.1$.} \label{fig:bickley-jet-overview}
    \squeezeup
\end{figure}
For this paper, the parameters are set to $\epsilon = 0.25, \alpha=0.25, A = 0.25, \text{and } \omega = 2 \pi$. We simulate $20\,000$ points uniformly sampled on grid $[0, 2] \times [0, 1]$. From time $t \in [0, 20]$ with step size $0.1$, we use a differential equations solver based on an explicit Runge-Kutta (4,5) formula on the velocities to calculate the trajectories of the points, as shown in Fig. \ref{fig:double-gyre-overview}.

With a Gaussian radial basis function kernel $k(x, x') = \exp(-(\norm{x - x'})/(2 \sigma ^2))$ for $\sigma = 0.75$, we compute the Gramian matrices for $G_{XX}$ for the initial position of the particles and $G_{Y_iY_i}$ for the position of the particles at time $i$, as in Eq. \eqref{time-dep-kcca}. In an online setting, only the inner sum needs to be recomputed as a moving average. The total product and its eigenvalue decomposition is performed. A $k$-means clustering for $k = 3$ is computed on the $3$ dominant eigenvalues of the estimated time-averaged matrix Eq. \eqref{time-avg-astar-a}.

\subsection{Bickley Jet}\label{subsec:bickley-jet}
The Bickley jet model is a prototypical model in the study of coherence that is a meandering zonal jet, flanked both above and below by counter rotating vertices. The Bickley jet is used as an idealized model for the Gulf Stream in the ocean and polar night jets in the atmosphere \cite{Del-Castillo-Negrete1992ChaoticFlow, Beron-Vera2010Invariant-tori-likeFlows}.

The stream function for the Bickley jet model is 
\begin{align*}
    \psi(x,y,t) &= \psi_0(y) + \psi_1(x,y,t) \\
    \psi_0(y) &= -U_0 L_0 \tanh\left(\frac{y}{L_0}\right) \\
    \psi_1(x, y, t) &= U_0 L_0 \sech^2\left(\frac{y}{L_0}\right) \Re \left({\sum_{n=1}^3 f_n(t) \exp(iknx)}\right), 
\end{align*}
with $f_n(t) = \epsilon_n \exp(-ik_nc_nt)$. The velocities can be computed as $\dot{x} = {\partial{\psi}}/{\partial{x}}$ and $\dot{y} = {\partial{\psi}}/{\partial{y}}$. We use scaled parameters from \cite{Hadjighasem2017ADetection} resulting in $U_0 = 5.4138, L_0 = 1.77, c_1 = 0.1446U_0, c_2 = 0.2053U_0, c_3 = 0.4561U_0, \epsilon_1 = 0.075, \epsilon_2 = 0.4, \epsilon_3 = 0.3, r_0 = 6.371, k_1 = 2/r_0, k_2 = 4/r_0, k_3 = 6/r_0$.

We sample $9900$ points uniformly on a grid $[0, 20] \times [-3, 3]$. From time $t \in [0, 40]$ with step size $0.1$, we calculate the trajectories of the points using a variable-step, variable-order Adams-Bashforth-Moulton solver of orders 1 to 13 on the velocity $[\dot{x}, \dot{y}]$, as shown in Fig. \ref{fig:bickley-jet-overview}. Some particles escape the grid $[0, 20] \times [-3, 3]$, and the positions of these particles wrap around such that they stay within the grid. The true time lag parameter $\tau$, as introduced in Sec. \ref{sec:online}, for this map is known to be $40$ for this simulation.

For this example, we also use a Gaussian kernel, with $\sigma = 1$. We compute the Gramian matrices for $G_{XX}$ for the initial position of the particles and $G_{Y_iY_i}$, as before for the time-varying Eq. \eqref{time-dep-kcca} and perform $k$-means clustering for $k = 9$  with $9$ dominant eigenvalues of the operator, similar to \cite{Klus2019KernelData}.

\subsection{Analysis of Online Coherent Sets Computation}
\begin{table}[!htb]
\squeezeup
\centering
\caption{Comparison of Evaluation Metrics on Methods for Computing Coherent Sets}
\renewcommand{\arraystretch}{1.3}
\begin{tabular}{|c|c||c|c|c|c|} \hline
\cline{1-5}
\multirow{2}{*}{\begin{tabular}{c}Environment \end{tabular}} & \multirow{2}{*}{\begin{tabular}{c}Method \end{tabular}} & \multicolumn{3}{c|}{Evaluation Metric} \\ \cline{3-5}
& & RI & H & V \\ \cline{3-5} \hhline{|=====|} 
\multirow{2}{*}{\begin{tabular}{c} Time-Dependent  \\  Double Gyre \end{tabular}} & \multirow{2}{*}{\begin{tabular}{c}Offline \\ Online\end{tabular}}  & 0.981 & 0.965 & 0.964  \\ \cline{2-5}
& & 0.967 & 0.940 & 0.945  \\ \hhline{|=====|}
\multirow{2}{*}{\begin{tabular}{c}Bickley Jet \end{tabular}} & \multirow{2}{*}{\begin{tabular}{c}Offline \\ Online \end{tabular}} & 0.696 & 0.559 & 0.628 \\ \cline{2-5}
& & 0.745 & 0.585 & 0.663 \\  \cline{1-5}
\end{tabular} \label{tab:cluster-eval}
\squeezeup
\end{table} 

We compare the online algorithm, using Eq. \eqref{time-dep-kcca}, to the offline algorithm, where we naively compute surrogate eigenvalues of the operator $\mathcal{A}^*\mathcal{A}$ at each snapshot. We compare methods for external cluster validation, where evaluation is based on comparing the online and offline clustering to a known true clustering. The true clustering is found by constructing the operator $\mathcal{A}^*\mathcal{A}$ for the true time parameter $\tau$ and using the $k$-means algorithm for the known number of clusters over multiple random initializations of cluster centers and averaging the clusters. The results are shown in Table \ref{tab:cluster-eval}. We use the Rand index adjusted for chance (RI), homogeneity (H), and the V-measure (V). The Rand index measures similarity of the assignments, homogeneity indicates if each cluster contains only members of a single class, and V-measure is a combination of homogeneity and completeness, where completeness indicates if all members of a given class are assigned to the same cluster. The evaluation metric is found by comparing clusters from each time step against the true clusters. This is then averaged across all time steps.
\begin{figure}[!htb]
    \captionsetup[subfloat]{farskip=2pt,captionskip=1pt}
    \centering
    \subfloat[ ]{\includegraphics[width=0.24\textwidth]{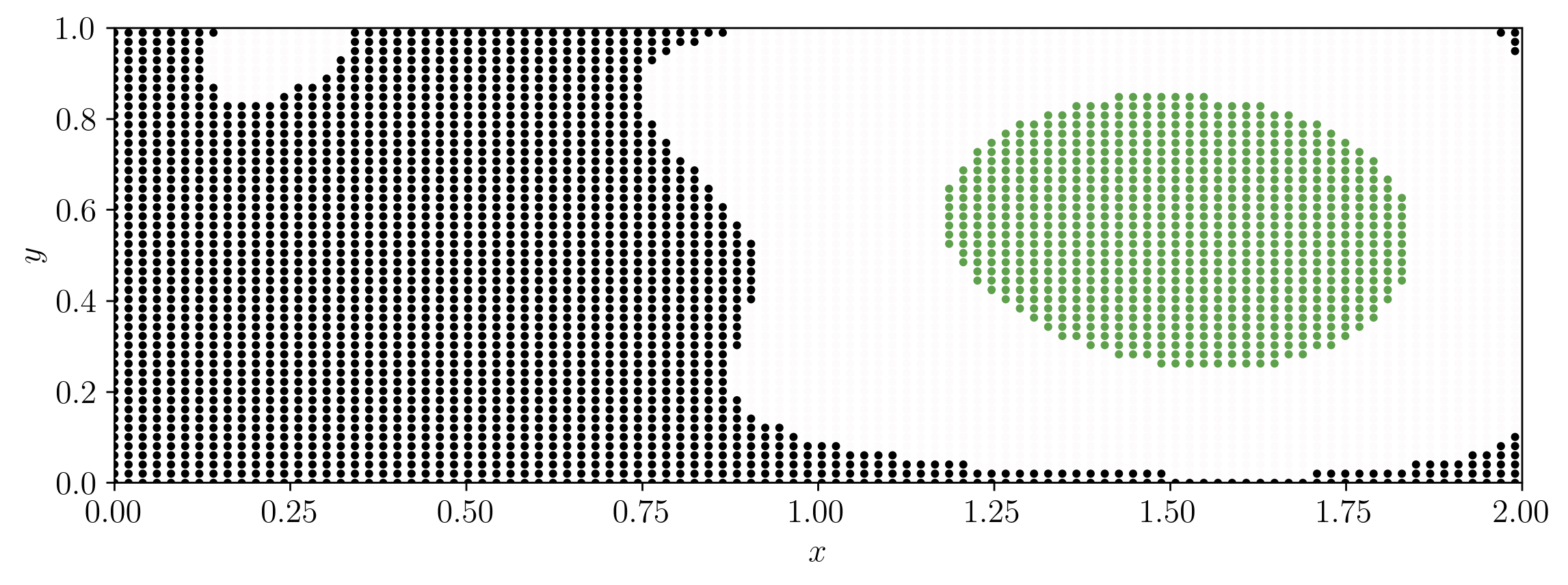}}
    \subfloat[ ]{\includegraphics[width=0.24\textwidth]{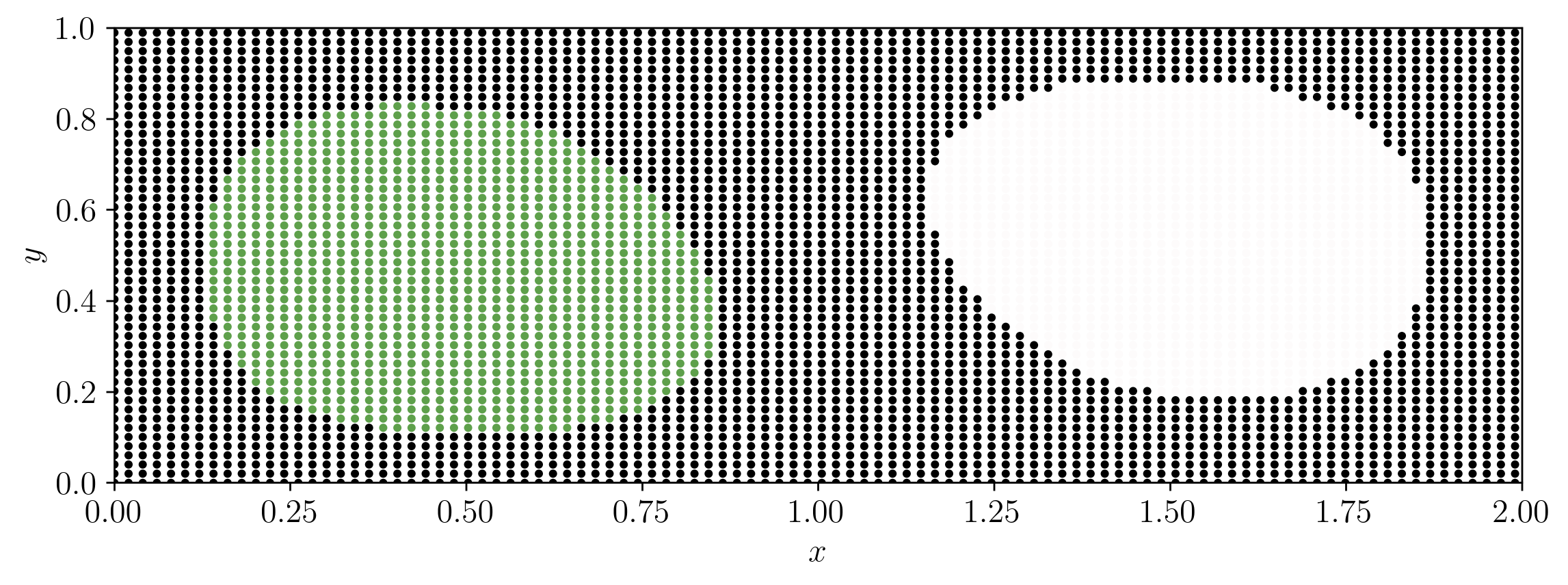}}
    
    \subfloat[ ]{\includegraphics[width=0.24\textwidth]{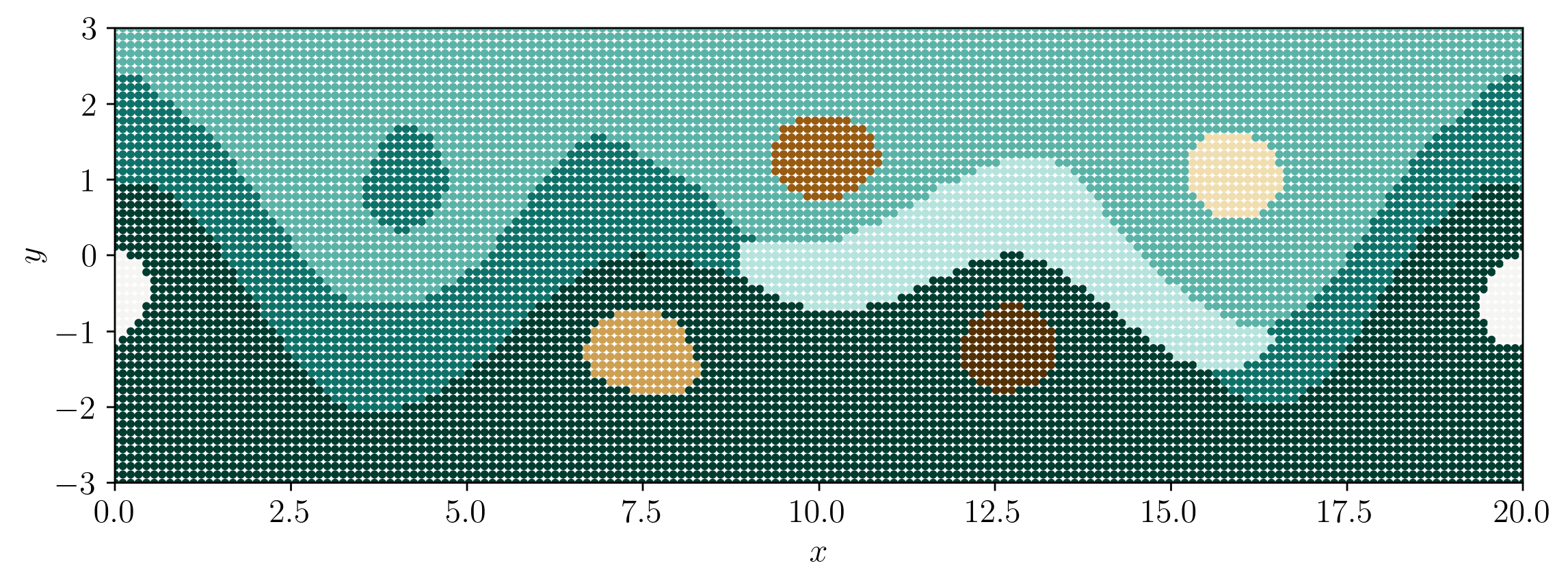}}
    \subfloat[ ]{\includegraphics[width=0.24\textwidth]{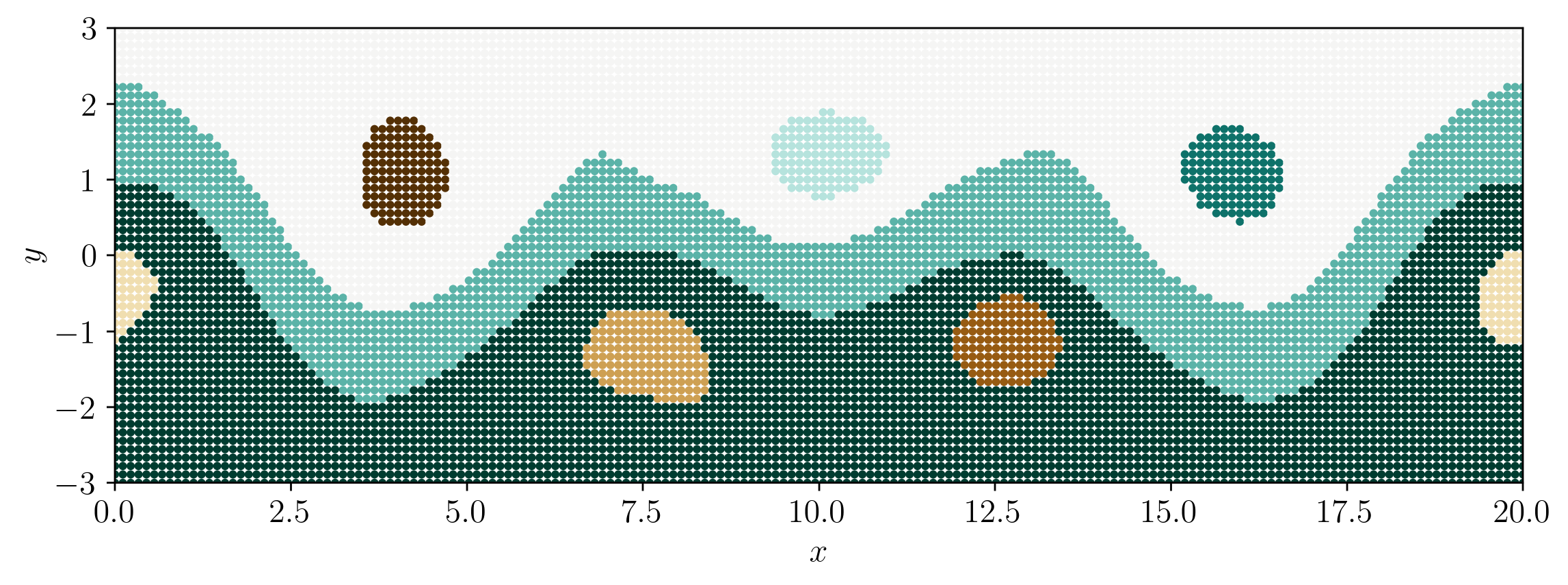}}
    \caption{Comparison of offline and online clustering algorithms for the time-dependent double gyre flow at time $t = 17.6$ in (a) and (b) and Bickley jet at time $t = 39.9$ in (c) and (d). (a) The offline version does not take into account prior measurements of the double gyre. (b) The online version incorporates historical data about the flow and consistently computes the true coherent sets. (c) The offline clustering of the Bickley jet incorrectly computes the coherent sets, even when close to the true time parameter. (d) The online method produces robust estimates of the coherent sets. The true coherent sets associated with the double gyre and Bickley jet almost exactly align with the coherent sets detected in (b) and (d).} \label{fig:dg-online-comp}
    \squeezeup
\end{figure}

Across all metrics, the online method for the double gyre performs slightly worse than the offline method.  In both the offline and online clustering, the clustering are qualitatively similar to the true clustering in size and shape. For the more complex and realistic Bickley jet, the online method outperforms the offline method across all metrics. In both scenarios, the online method is less sensitive to noise and invalid clustering, as shown in Fig. \ref{fig:dg-online-comp}. The simulations and experiments conducted in this paper rely on the online method, which requires less knowledge about the environment and works better in complicated, real world scenarios due to its incorporation of data over time.

\section{Environmental Feature Generation for Aerial and Surface Robots  \protect\footnote{Accompanying video can be found at \href{https://youtu.be/D-4S6hnFr2E}{https://youtu.be/D-4S6hnFr2E}}}
In this section, we discuss how computing coherent sets in real time facilitates robots' awareness. We then demonstrate preliminary ideas as to how this information can be incorporated into the robots' decision-making. 

\subsection{Aerial vehicle based crowd monitoring}\label{sec:av-crowd}
Robots operating in dense environments, such as urban landscapes or roadways, rely on persistent monitoring of crowds for scene understanding and safe navigation. Much of the literature in this area surrounds tracking individual trajectories of pedestrians or other objects. However, understanding global information about pedestrian movement allows for greater awareness of high-level behaviors and facilitates informed planning. With aerial vehicles monitoring an environment overhead, we can collect trajectory information of pedestrians, which can be thought of as a flow-like environment \cite{Ali2007AAnalysis}. This information can be used within our framework to elucidate key features, such a high congregation areas, of the underlying, persistent crowd behaviors.

Overhead data of pedestrians at a train station stop was collected from the $4$th floor of an hotel in Bahnhofstrasse, Zurich \cite{pellegrini2009you}. The original sequence is processed with a time step of $0.4 s$. We found and analyzed the dataset using tools from OpenTraj \cite{amirian2020opentraj}. We compute coherent sets using trajectories of $116$ pedestrians over $51$ time steps, with a $2$-degree polynomial kernel and $3$-means clusters on $3$ eigenvalues.

\begin{figure}[!htb]
    \centering{\includegraphics[width=0.46\textwidth]{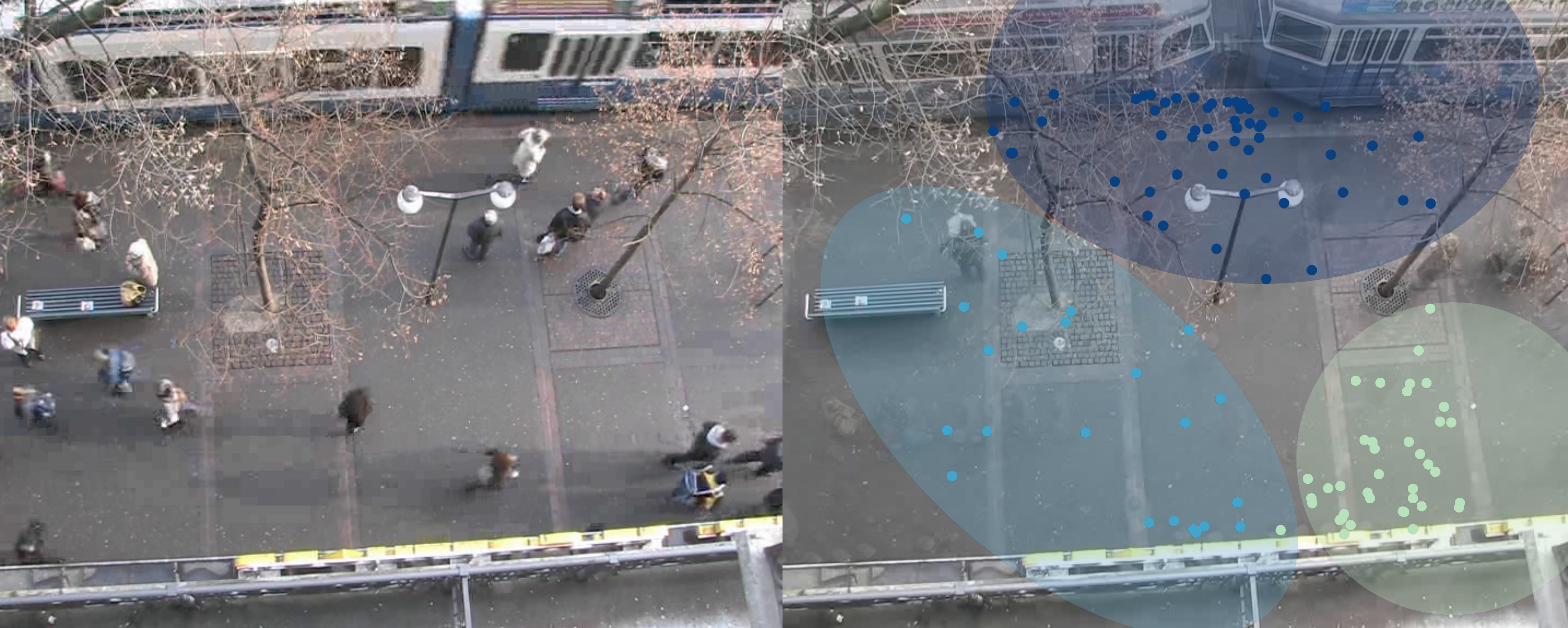}}
    \caption{Aerial view of pedestrians walking through train station with coherent sets overlaid. Coherent sets correspond to regions of interest including train platform, entry areas, and exit areas, indicated by different colored circles. Regions could be used for decision-making in urban or dense environments.}  \label{fig:monitoring-cs}
\end{figure}

In Fig. \ref{fig:monitoring-cs}, we see coherent sets illustrate areas of congregation or high movement for pedestrians. Pedestrian data can be collected in real time to track coherent sets online, where sets would then be used to navigate through crowds, such as by avoiding the busy train platform, or deploy sensors for situational awareness, such as by monitoring the entrance for anomalous behavior or notable changes in coherence patterns. The regions of interest are areas where if groups of pedestrians start within a specified region, they tend to exhibit similar behaviors. The proposed indications of various areas of interest, such as regions of waiting, entry, and exit, provide a semantic interpretation of the coherent sets in this specific context.

\subsection{Energy-efficient trajectories for surface vehicles}
Currents in rivers, oceans, and in the atmosphere are  other examples of flow fields. While global information is complex and hard to analyze, local information does not provide an accurate description of environment features. Here, we leverage coherent sets from a single gyre to demonstrate the usefulness of path planning under awareness of these global features.

\begin{figure}[!b]
    \centering{\includegraphics[width=0.46\textwidth]{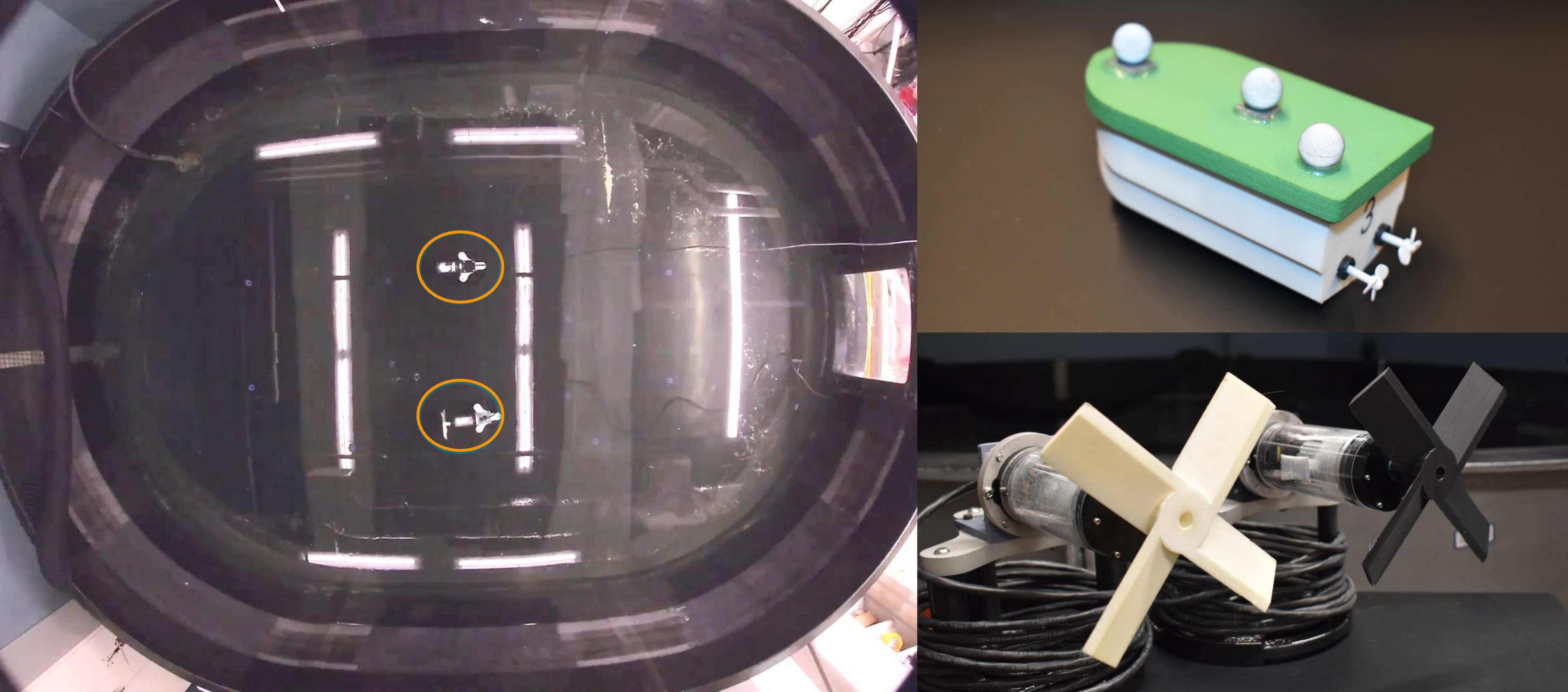}}
    \caption{Experimental test bed for surface vehicle navigation in controlled flows within a tank. The mCoSTe consists of a mASV and a water tank, where two flow driving cylinders were placed horizontally in the tank, circled in orange, to create a single gyre.} \label{fig:boat-setup}
\end{figure}

We create flows and deploy surface vehicles using the multirobot Coherent Structure Testbed (mCoSTe) which consists of a micro Autonomous Surface Vehicle (mASV) and a Multi Robot flow tank (MR tank). The MR tank is a $4.5m \times 3.0m \times 1.2m$ tank of water equipped with an OptiTrack motion capture system providing localization of the mASV at $120Hz$. The mASV is a differential drive vehicle with a maximum forward speed of approximately $0.2 m/s$. Two flow driving cylinders placed horizontally and rotating at approximately $200rpm$ creates a single gyre flow in the MR tank, as shown in Fig \ref{fig:boat-setup}. 

\begin{figure}[!ht]
\centering
\includegraphics[width=0.48\textwidth]{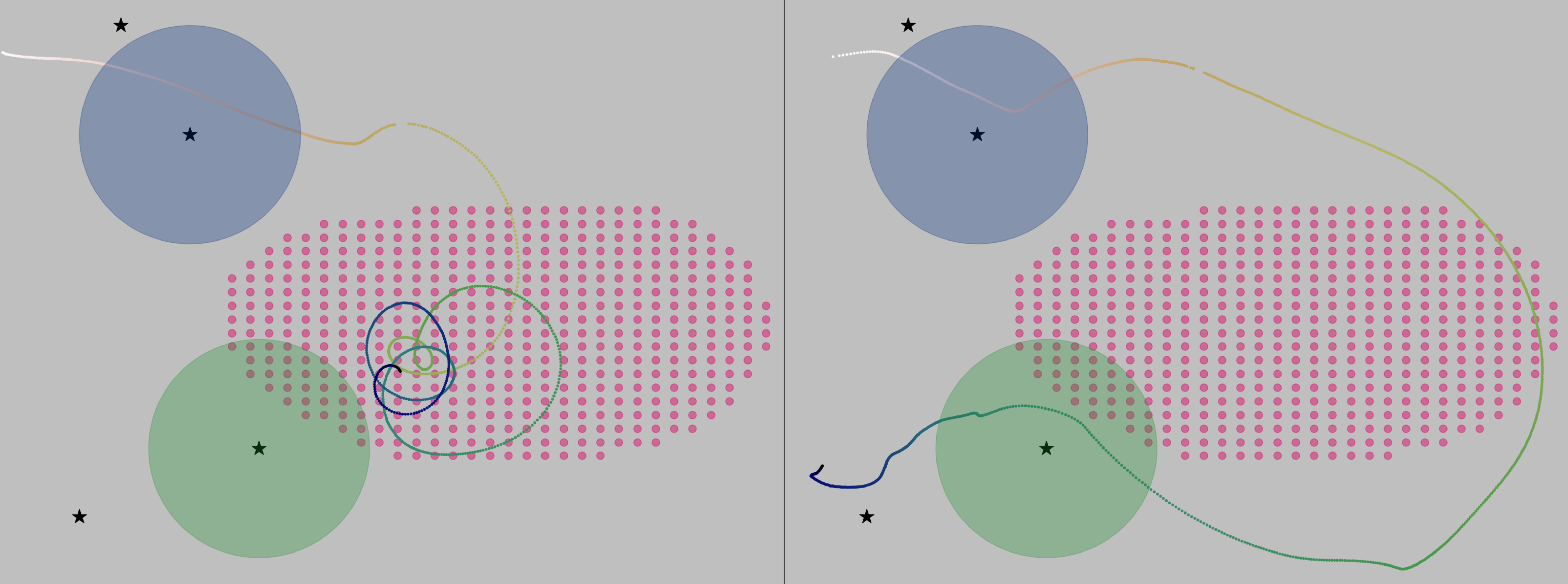}
\caption{Comparing trajectories of a robot operating in a single gyre flow and using a coherent set in decision-making, with region of entry (blue), region of exit (green), coherent set (pink), and selected waypoints (stars). The trajectory of the mASV is lightest where it began and darkest where it ends. (Left) The trajectory of an mASV stuck in the single gyre while attempting to reach the goal. (Right) The trajectory of an mASV leveraging the coherent set to move through the environment efficiently and escape when close to the goal.}
\label{fig:boat_traj_data}
\squeezeup
\end{figure}

We assume one or more robots are collecting velocity or trajectory data. This may be aerial vehicles collecting overhead imagery of the flow field or passive gliders directly sampling the velocity. This information is then processed in real time to compute the coherent sets. Instead of surface vehicles tracking and modeling individual trajectories within the flow field, the robot maintains the coherent sets as a global description of the dynamics of its environment. The coherent sets are calculated using an approximate flow model for the gyres in the tank for $3591$ simulated particles over $150$ time steps, where the flow model is a noisy estimate for the true flow induced in the tank by the flow driving cylinders. Our coherent set algorithm uses a Gaussian kernel with $\sigma = 1.25$ and $3$-means clusters on $3$ eigenvalues. Then, the sets are used to construct energy-efficient paths. We plan paths from a specified start location to a goal location by picking waypoints near coherent sets, exerting control effort to travel to and escape from coherent sets, and remaining in coherent sets until the next waypoint is close. This is ideologically similar to the waypoint selection method from \cite{Ramos2018LagrangianMissions} using LCS.

The robot navigates the flow induced by a single gyre model and completes its mission when planning a trajectory based on the coherent set. During execution, the robot saves energy by turning off its motors when the boundary of the coherent set is reached and floating through the minimal dispersion region. However, a boat unaware of the coherent set drives into the gyre, causing the robot to become stuck and ultimately fail to reach its goal, as seen in Fig. \ref{fig:boat_traj_data}. Modeling coherent sets gives the robot a descriptive framework to reason about flows in the environment, without maintaining detailed knowledge. 


\section{Conclusion}
Robots operating in complex scenarios necessitates greater awareness of their environment. In flow-like environments, global dynamics may be more useful than tracking individual trajectories. Transfer operator theory provides a systematic framework for reasoning about global dynamics. Recent advances in machine learning and transfer operators combine powerful data representation with physical system knowledge.  Here, we present a framework where coherent sets, defined by transfer operators, are learned with kernel methods, estimated online, and integrated into decision-making for robots. We foresee robotics applications incorporating awareness of environments in novel, meaningful ways with these connections. For example, our framework allows scene segmentation according to pedestrian movement instead of static objects and efficient navigation of strong flows with little information. Future work includes automated kernel function selection, a distributed framework for sensor information collected by multi-robot teams, development of more sophisticated decision-making schemes, and analysis in real environments. 

\bibliographystyle{IEEEtran}
\bibliography{references}


\end{document}